# DEEP TRANSFORMER NETWORKS FOR TIME SERIES CLASSIFICATION: THE NPP SAFETY CASE

**Bing Zha[1], Alessandro Vanni[3], Yassin Hassan[3], Tunc Aldemir[2] and Alper Yilmaz[1]**

[1]Photogrammetric Computer Vision Lab, The Ohio State University
2070 Neil Avenue, Columbus, OH, USA
zha.44@osu.edu; yilmaz.15@osu.edu

[2]Nuclear Engineering Program, The Ohio State University
201 W. 19th Avenue, Columbus, OH, USA
aldemir.1@osu.edu

[3]Department of Nuclear Engineering, Texas A&M University
Building 3133, 423 Spence St, College Station, TX, USA
av_ale@tamu.edu; y-hassan@tamu.edu

## ABSTRACT

A challenging part of dynamic probabilistic risk assessment for nuclear power plants is the need for large amounts of temporal simulations given various initiating events and branching conditions from which representative feature extraction becomes complicated for subsequent applications. Artificial Intelligence techniques have been shown to be powerful tools in time dependent sequential data processing to automatically extract and yield complex features from large data. An advanced temporal neural network referred to as the Transformer is used within a supervised learning fashion to model the time dependent nuclear power plant (NPP) simulation data and to infer whether a given sequence of events leads to core damage or not. The training and testing datasets for the Transformer are obtained by running 10,000 RELAP5-3D NPP station blackout scenarios with the list of variables obtained from the RAVEN software. Each simulation is classified as 'OK' or 'CORE DAMAGE' based on the consequence. The results show that the Transformer can learn the characteristics of the sequential data and yield promising performance with approximately 99% classification accuracy for the testing dataset.

*Key Words: NPP Safety, Time Series, Deep Learning, Transformer Network*

## 1    INTRODUCTION

Safety is the utmost important requirement in nuclear power plant (NPP) operation. However, NPPs are highly complex system monitored by human operators and the large amount of high-dimensional data generated by dynamic probabilistic risk assessment (DPRA) tools [1] makes it hard to visually infer the consequences of events occurring during a NPP accident scenario, equipment failure or an external disturbance to the system. If core damage occurs at an NPP, the operating staff in the control room is responsible for returning the plant to a safe stable state. They are supported in taking certain actions, in particular, implementing emergency operating procedures for which they receive extensive training in plant simulators. Most of these actions are carried out by personnel who has qualifications and the training to perform their responsibilities which are heavily dependent on the experience of the person. However, the ability of plant management to project the consequence of an event plant that is in the early stages of development is currently limited. Therefore, it is important to develop a framework in which the abnormal condition for a given event can be identified on time to give the plant operators adequate time to perform corrective actions.

The identification of the possible consequence of an initiating event such as station blackout (SBO) can be considered a pattern recognition problem. When the event occurs starting from steady-state operation, a simulator can be used to develop a time-dependent pattern that is unique to a specific sequence and identify an abnormal consequence. However, simulation of accidents can take days, producing terabytes of data [2]. In situations with the availability of such massive data and powerful computational resources, artificial neural networks (ANN) have been shown in many occasions to be the best pattern recognition tool for early identification of consequences. Various ANN models such as convolutional neural networks (CNNs) [3] and recurrent neural networks (RNNs) [4] have been studied and also successfully applied on nuclear data [5, 6, 7]. This paper introduces a more advanced sequence model called Transformer network [8] that combines the benefits of CNNs and RNNs and has been shown to obtain significantly better performance than prior ANN models in the field of machine learning.

The objective of this paper is to present a framework using dynamic event trees (DETs) and deep learning (DL) technique for early identification of the consequences of a sequence of events for NPP safety analysis. DETs were used to simulate a time-dependent flow of information that accounts for the uncertainties associated with the severe accident progression. DL is employed to automatically extract features from NPP simulator data.

The remainder of this paper is structured as follows: we provide background on the system used in generating dataset and the artificial intelligence (AI) technique used in Section 2 for ANN construction. The methodology is presented in Section 3. Experiments and results are shown in Section 4. Finally, we conclude the paper in Section 5.

## 2   BACKGROUND

A brief overview of the system and initiating event with possible branching conditions under consideration are given in Section 2.1. Section 2.2 provides background on AI and DL.

### 2.1 System Overview

The initiating event under consideration is a SBO in a 4- loop pressurized water reactor (PWR). The evolutions of the SBO that is considered for this study includes the loss of coolant accident such as the one caused by the failure of the reactor coolant pumps seals or by the pressure operated relief value (PORV) failing to reclose.

The possible scenarios (total of 10,000) following the initiating event were generated using RAVEN [9] and RELAP5-3D [10]. Table I shows the major RELAP5 -3D modeling parameters. Figure. 1 shows the event tree (ET) structure used for scenario generation.

**Table I. Major RELAP5-3D Modeling Parameters**

| Parameter | Unit | Value |
|---|---|---|
| Power Plant Type | - | Typical 4 loop PWR |
| Core Thermal Power | [MWth] | 3850 |
| Total Primary Volume | [ft3] | 14300 |
| Secondary Volume[1] | [ft3] | 30500 |
| Steam Generators PORV Setup Point | [psia] | 1235 |
| Pressurizer PORV Opening Setup Point | [psia] | 2350 |
| Pressurizer PORV Closing Setup Point | [psia] | 2330 |
| Number of TDP AFW | - | 1 |
| AFW Battery Depletion Time | [h] | 4 |

TDP: Turbine Driven Pump    AFW: Auxiliary Feedwater

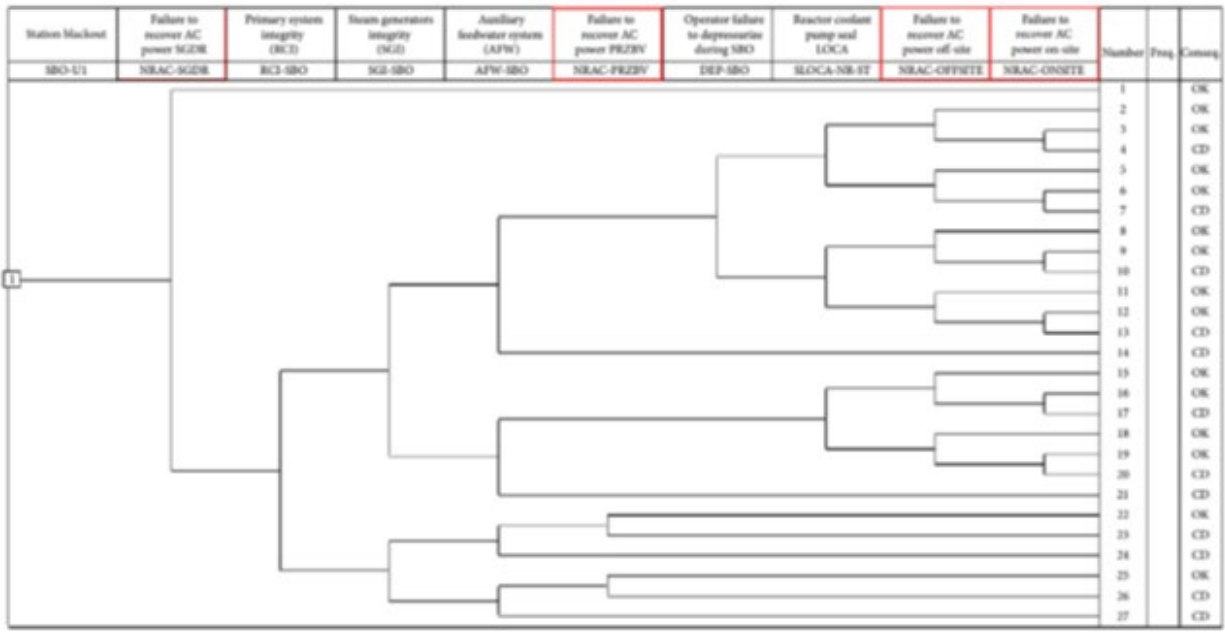

**Figure 1. ET structure used for scenario generation.**

## 2.2 Deep Learning Overview

The field of AI officially started in 1956, launched at a small but now famous summer conference at Dartmouth College. Simply put, AI is a broad concept that refers to the ability of a machine to mimic the capabilities of the human mind by learning from examples and experience. The evolution of AI has gone through three significant waves. The first wave of AI begun in the 1950s and was based around "handcrafted" knowledge with which system could use logic to find specific solutions. The transition from the first to the second wave of AI happened around 2000s, during which the distinguishing element was the application of statistical methods. The third wave of AI uses machine learning to automatically learn from vast amounts of data and improve from experience without being explicitly programmed with knowledge. Today, AI is a thriving field with many practical applications ranging from engineering, science and medicine.

DL is an approach within AI that allows a machine to be fed with raw data to automatically discover the representations needed for specific problems by composing multiple ANNs that transform the learned representation at one level into a representation at a higher level [11] as shown in Fig. 2. The DL technique has shown great success in time-dependent continuous data, such as video, audio and text which makes it suitable for analyzing temporal characteristics of event evolution within an NPP. There are two common neural network architecture used in DL called CNNs and RNNs. The CNNs is primarily used with grid-like data structure, such as an image, while RNNs are designed for processing data exhibiting temporal dynamic behavior, like a video and audio. The recent proposed novel network architecture Transformer used in this paper enables the parallelization and has a better performance.

---
[1] Considering only the volume of secondary side the four steam generators

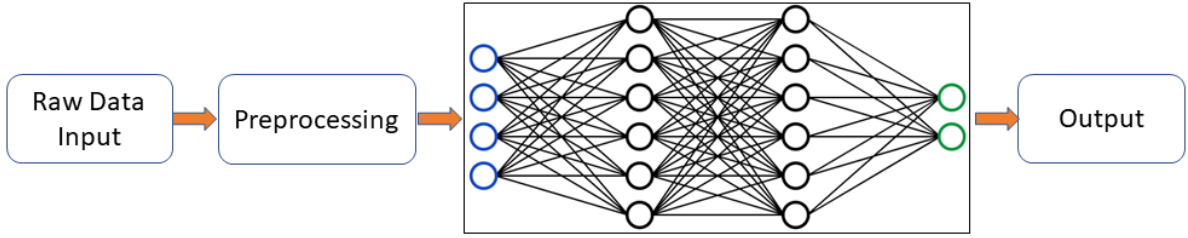

**Figure 2. The typical deep learning working pipeline using ANN where the circle represents "neuron" and the line connection between "neuron" is called weight parameter which is exactly needed to be learned. Each column of "neurons" is named "layer".**

## 3  METHODOLOGY

### 3.1 Data Representation

Each branch of the ET in Fig. 1 representing a scenario contains time evolution of a large number of state variables and thus the resulting data are high-dimensional in both state variables and time. To describe the temporal behavior of all system state variables (e.g., pressure and temperature), we represent each scenario $x_i (i = 1, ..., I)$ by $M$ state variables $x_{im}(m = 1, ..., M)$ and time length $T$ as the $M * L$ matrix:

$$x_i = \begin{bmatrix} x_{i1}(t_1) & \cdots & x_{i1}(t_L) \\ \vdots & \ddots & \vdots \\ x_{iM}(t_1) & \cdots & x_{iM}(t_L) \end{bmatrix} \quad (1)$$

where, $x_{im}(t_l)$ corresponds to the value of the variable $x_m$ sampled at time $t_k$ (e.g., $t_1 = 0$ and $t_L = T$) for scenario $i$. Note that the dimensionality of can be extremely high when state variables and time interval become large. In this work, a smaller set of variables of interests are chosen manually by expert judgement which will be discussed in Section 4.

Another issue that arises when dealing with nuclear transients due to different scale of the variables used. Hence, each variable in scenario needs to be normalized into the range of 0 and 1. The normalization procedure is defined as:

$$\bar{x}_{im} = \frac{x_{im} - min(x_{im})}{max(x_{im}) - min(x_{im})} \quad (2)$$

where, $min(x_{im})$ and $max(x_{im})$ are the respective minimum and maximum of the variable of the scenario $i$ in the entire time length.

We show two scenario examples after normalization in Fig. 3.

### 3.2 Transformer Network

The Transformer network [8] was initially proposed for machine translation problem where the input and output are both sequence data, but due to its high performance in processing sequential data it was quickly incorporated into other areas, such as music [12] and earthquake waves [13]. Transformer network

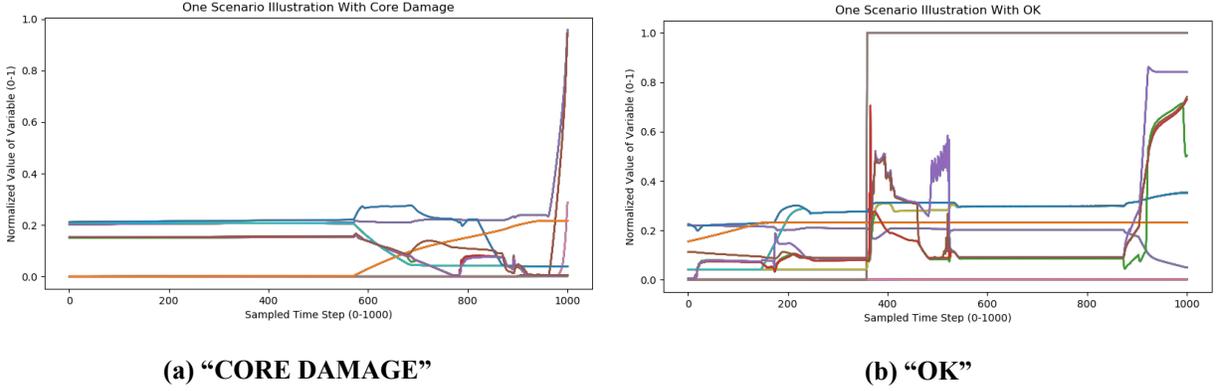

**(a) "CORE DAMAGE"**  **(b) "OK"**

**Figure 3.** Visual illustration of two different scenarios with "CORE DAMAGE" and "OK" case where each color indicates the evolution of different variable.

uses an attention mechanism. The fundamental idea of attention is to learn a scoring function that assigns a different weight to each piece of given data. Specifically, input temporal data $x_i \in R^{L \times M}$ of length $L$ and dimension $M$ are firstly linearly projected into a set of keys $K \in R^{L \times d}$, queries $Q \in R^{L \times d}$ and $V \in R^{L \times d}$ where $d$ is the dimension of $K, Q$ and $V$. With those key-value element pairs $(k_i, v_i)$ and a query $q$, an attention model computes weights of each key with respect to query, and aggregates the values with these weights to form the value corresponding to the query as defined by:

$$Attention(Q, K, V) = softmax\left(\frac{QK^T}{\sqrt{d}}\right)V. \qquad (3)$$

where, the *softmax* function in Eq. (3) is used to normalize an input value into a vector of values that follows a probability distribution whose total sums up to 1. In Eq. (3), self-attention [8] is achieved when the keys $K$ and queries $Q$ are identical. The main benefits of this mechanism over RNN and CNN are the ability to capture long dependencies between input data and at the same time train neural network in parallel.

In this work, we only adopt encoder part of Transformer since our goal is to classify the temporal data instead of generating another sequence. As shown in Fig. 4, the Transformer encoder is composed of an input layer, a positional encoding layer and a stack of $N$ identical encoder layers. The input layer maps the input time series data to a vector of another dimension through a fully-connected (FC) network which is an essential step for the model to employ attention mechanism. Since there is no information of the order of the sequence due to the fact that the model contains no recurrence and convolution, the positional encoding is hence used to indicate the relative or absolute position of the input sequence data by adapting a sine and cosine functions of different frequencies to each input element. The resulting vector is fed into N encoder layers where N is pre-defined and indicates the depth of the neural network. Each encoder layer consists of two sub-layers: self-attention and fully-connected feed-forward layer with following addition and normalization operation. The whole architecture transforms the input sequence into another new output sequence which has incorporated information from every other input element.

### 3.3 Temporal Data Classification

The goal of this paper is to predict the consequences of each scenario based on its input temporal data. The overall pipeline is illustrated in Fig. 5. First, the preprocessed multivariate time series data is fed into Transformer encoder network as a feature extractor, resulting in a new representation of input data. Then, the FC and *softmax* function as a classifier are utilized to predict whether the input scenario data will cause "CORE DAMAGE" or be "OK".

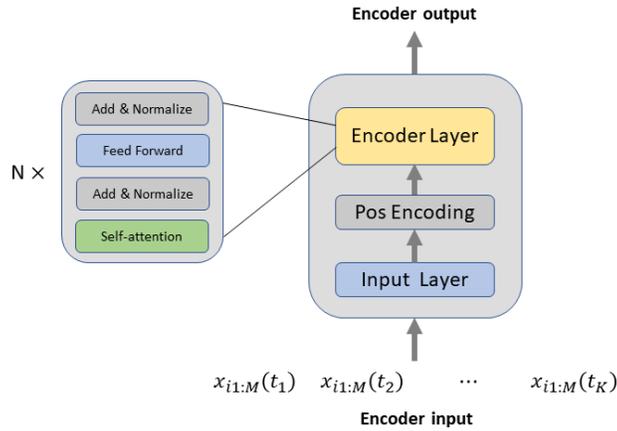

**Figure 4. Architecture of Transformer-based encoder model for representing multivariate time series data.**

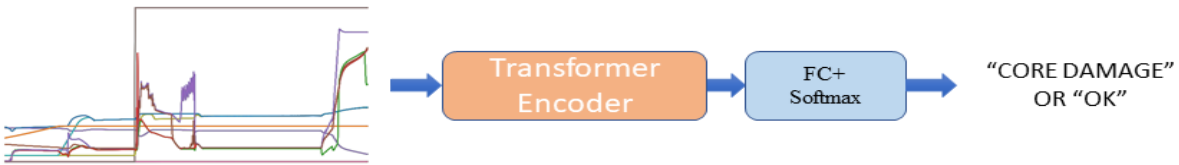

**Figure 5. Overall of the pipeline to classify each scenario.**

## 4    EXPERIMENTS AND RESULTS

### 4.1  Dataset Overview

The scenarios dataset are generated from selected initiating events and branching conditions provided by Texas A & M University for 10,000 RELAP5-3D SBO simulations with the list of RAVEN variables given in Table II. For the training purpose, each scenario is associated with a corresponding label. Here, the scenario is labelled as "OK" for each case that the final value of "maxPCTdegF" is smaller than 2100, while labelled as "CORE DAMAGE" for any other cases. The total number of scenarios is 9,587.

For the purpose of application in machine learning, the dataset is partitioned into three subsets: training, testing and validation as shown in Table. III. The training dataset accounting for 64% dataset is used to fit the parameters of the neural network. The validation set is a separate section of the dataset that is employed to get a sense of how well the model is performing. The testing dataset corresponds to the final evaluation after all of the training experiments have concluded.

**Table II. List of input RAVEN variables**

| Variable Name | Variable Description |
|---|---|
| PCTdegK | PCT degK |
| maxPCTdegK | Maximum of PCT in degK |
| PCTdegF | PCT in degF |
| maxPCTdegF | Maximum of PCT degF |
| totGeneratedHydrogen | Total (cumulative) generated hydrogen from the mental-water reaction model (kg) |
| cntrlvar_601 | Availability of the power from the grid |
| SG2CoolantInventory | Secondary side of SG2 coolant inventory |
| SG3CoolantInventory | Secondary side of SG3 coolant inventory |
| SG4CoolantInventory | Secondary side of SG4 coolant inventory |
| i_volflowSRCP1LOCA | Cumulative volumetric follow rate from small RCP LOCA break (loop 1) [GPM] |
| subcoolingCL1 | subcooling level in the cold leg 1 [K] |
| subcoolingCL2 | subcooling level in the cold leg 2 [K] |
| subcoolingCL3 | subcooling level in the cold leg 3 [K] |
| subcoolingCL4 | subcooling level in the cold leg 4 [K] |

PCT: Peak Clad Temperature SG: Steam Generator LOCA: Loss of Coolant Accident
RCP: Reactor Coolant Pump

**Table III. Number of REALAP-3D runs used for training, testing and validation**

| | CORE DAMAGE (0) | OK (1) | TOTAL |
|---|---|---|---|
| **TRAINING** | 3775 | 2345 | 6192 (64%) |
| **TESTING** | 1190 | 726 | 1916 (20%) |
| **VALIDATION** | 941 | 592 | 1533 (16%) |
| | | | |
| **TOTAL** | 5906 | 3672 | 9587 (100%) |

### 4.2 Implementation Details and Training Process

The Transformer network is implemented using open-source AI framework PyTorch [18] which contains various architectures of ANNs and which can be easily utilized with input and output data. The training process is completed on a computer with NVIDIA GTX1080 GPU. Considering that predicting the consequences is essentially a binary classification problem, the cross-entropy function defined in Eq. (4) is utilized as the loss function for measuring the training performance:

$$Loss = -(ylog(p) + (1-y)\log(1-p)) \qquad (4)$$

where, $y$ is binary indicator (0 or 1) if class label is the correct classification for input data and $p$ is the predicted probability of outcome.

Training a neural network involves setting a number of hyperparameters. For our Transformer network, we use two layers and set hidden size as 30. To train the network, we use stochastic gradient descent [14] as an optimization method with a fixed learning rate 0.001 which can control the speed of weight update. Since we observe that the loss is quickly converged, the entire dataset is passed through the neural network only once.

### 4.3 Results and Analyses

The training results is shown in Fig. 6 where the loss can be observed to be converged quickly. To compare the training performance of different neural network architectures, three common networks are trained for comparison. The Transformer network and CNN converge faster than RNN at the beginning. While the final performance of three networks appears similar, we should note that as we get to 100% accuracy small increments show significant performance boost. By investigating the pattern of the two classes of scenarios as illustrated in Fig. 3, we have found out that they are significantly distinguishable with time sequence increasing which also indicates the scenario leading to the consequences of CORE DAMAGE or OK is easy to classify, and all three networks are able to extract representative features from scenario dataset. Nonetheless, the Transformer network is still the best choice to handle multivariate scenario time series data. First, the length of temporal data more than 1000 can be still processed. Second, the Transformer network can be implemented in parallel for larger dataset. The performance measurement we adopt is classification accuracy defined in Eq. (5) with results shown in Table IV.

$$Accuracy = \frac{\#\ true\ prediction}{\#\ total\ prediction} \quad (5)$$

**Table IV. Accuracy on training, testing and validation dataset with three different models**

|  | Model | Train | Valid | Test |
|---|---|---|---|---|
| **Accuracy** | Transformer | **99.6%** | **99.1%** | **99.0%** |
|  | CNN | 99.0% | 99.0% | 98.9% |
|  | RNN | 99.2% | 98.9% | 99.0% |

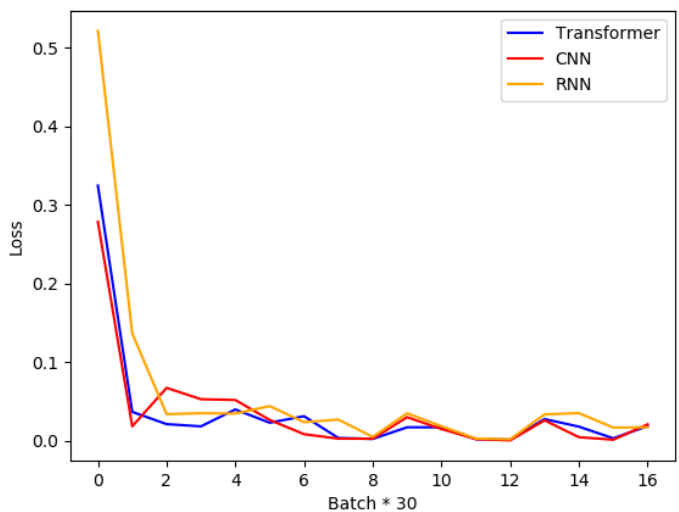

Figure 6. The training loss comparison with three different architectures. The "Batch" is the number of training examples used in one iteration and set as 12 here.

## 5    CONCLUSIONS

In this paper, we propose to use Transformer network structure to model multivariate time series data in the context of NPP safety analysis. The temporal data obtained from NPPs monitors are first preprocessed and then fed into Transformer network to classify whether the scenario will cause CORE DAMAGE or not. The experimental results show that Transformer network can obtain good performance and possess benefits over CNN and RNN.

## 6    ACKNOWLEDGEMENT

The research presented in this paper was partially supported by the U.S. Department of Energy Nuclear Energy University Program (NEUP) award 17-12723.